# Fre-CW: Targeted Attack on Time Series Forecasting using Frequency Domain Loss

Naifu Feng, Lixing Chen, Junhua Tang, Hua Ding, Jianhua Li, Yang Bai
Shanghai Jiaotong University, Shanghai, China
`{fnf112358,lxchen,junhuatang,dinghua623,lijh888,`
`ybai}@sjtu.edu.cn`

**Abstract.** Transformer-based models have made significant progress in time series forecasting. However, a key limitation of deep learning models is their susceptibility to adversarial attacks, which has not been studied enough in the context of time series prediction. In contrast to areas such as computer vision, where adversarial robustness has been extensively studied, frequency domain features of time series data play an important role in the prediction task but have not been sufficiently explored in terms of adversarial attacks. This paper proposes a time series prediction attack algorithm based on frequency domain loss. Specifically, we adapt an attack method originally designed for classification tasks to the prediction field and optimize the adversarial samples using both time-domain and frequency-domain losses. To the best of our knowledge, there is no relevant research on using frequency information for time-series adversarial attacks. Our experimental results show that these current time series prediction models are vulnerable to adversarial attacks, and our approach achieves excellent performance on major time series forecasting datasets.

## 1 Introduction

Research on time series forecasting has grown rapidly in recent years and has been widely used in areas such as stock market [1], energy consumption [2], and traffic flow [3]. With the development of deep learning (DL), transformer-based models are gradually becoming mainstream in the field. However, deep learning models are known to be vulnerable to adversarial attacks, where some small perturbations can cause the model to produce incorrect outputs [4]. Adversarial attacks on time series forecasts have more serious real-world consequences than attacks in other areas. Although adversarial attacks study time series forecasting models from an adversarial perspective, they can provide valuable insights for evaluating and enhancing the robustness of these models.

Adversarial attacks were first proposed in image classification and subsequently applied to time series classification tasks. However, there is little research on adversarial attacks against time series prediction models. One challenge to adversarial attacks against time series forecasting models comes from the nature of time series data. Unlike images, time series data typically have lower dimensions and degrees of freedom (e.g., window length). This limitation not only increases the likelihood of detection but also poses a significant challenge in designing attacks that can effectively fool prediction models while maintaining the realism and integrity of the data. In addition, time series data often has a fixed pattern (e.g., periodicity), and random perturbations may break the original data pattern, resulting in the exposure of an attack.

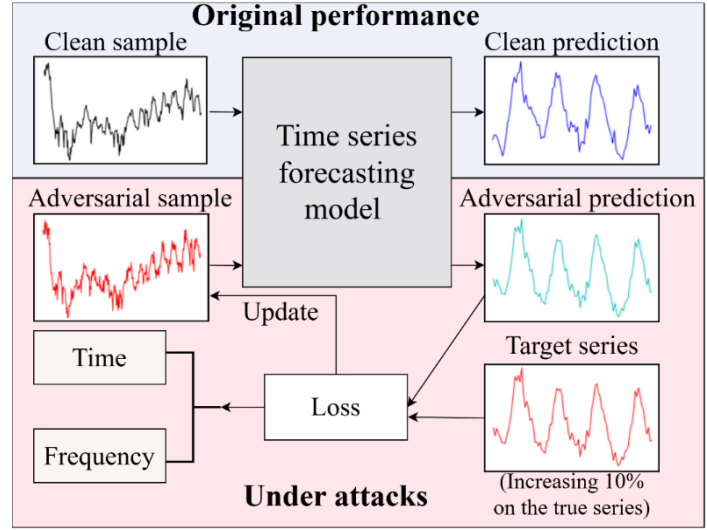

**Fig. 1.** Visualization of targeted adversarial attacks on time series forecasting.

Adversarial attacks on time series forecasting can be categorized into non-targeted attacks and targeted attacks [5,6]. Non-targeted attacks aim to make the prediction results significantly different from the true series, while targeted attacks aim to make the prediction results close to a specific sequence desired by the attacker. In terms of task characteristics, unlike the classification task, the optimization objective of the prediction task is no longer a clear classification hyperplane, but rather for the model output to be as close or as close as possible to a particular sequence. This means that each data point in the sequence varies based on the direction of its gradient, complicating the optimization of adversarial samples. Current algorithms for time-series prediction attacks are largely derived from image classification methods, often generating perturbations directly using gradients. In contrast, methods like the Carlini and Wagner adversarial attack mechanism(C&W) [18], developed in the image domain, directly optimize adversarial samples. These approaches generally surpass traditional gradient-based attacks, as demonstrated in the image domain, but have yet to be explored in the context of time series forecasting.

The importance of time series frequency domain information has been progressively studied. During the training process of forecasting models, the models usually focus more on capturing low-frequency components (e.g., trends and seasonality) and under-represent high-frequency components (e.g., abrupt changes). This is instructive for designing adversarial attacks. Adversarial perturbations with high-frequency noise can disrupt the model's ability to capture underlying temporal patterns, resulting in inaccurate predictions. However, blindly introducing high-frequency perturbations may also expose the attack, as they usually destroy the natural smoothness or periodicity of the original time series.

In this paper, we propose the Frequency-enhanced C&W Attack (Fre-CW), an adversarial attack method for long-time series forecasting models. The core idea of this approach is to simultaneously optimize the time and frequency domain losses of adversarial sequences. In addition, the attack algorithm designed for classification is adapted to apply to forecasting tasks.

The contributions of our work are as follows:

1. We have modified the C&W attack algorithm for time series forecasting attacks and replaced the misclassification function in the original algorithm with time series forecasting loss, which allows the C&W algorithm to be used in non-targeted and targeted attacks.

2. We propose the Frequency-enhanced C&W Attack (Fre-CW), a simple but effective modification of the C&W attack algorithm by introducing frequency domain loss to generate adversarial samples. To the best of our knowledge, this is the first work to use frequency analysis to enhance time series adversarial attacks.

3. We conduct experiments on six widely used datasets for two mainstream time series prediction models and demonstrate that Fre-CW achieves state-of-the-art attack performance.

## 2    Related Work

### 2.1    Multivariate time series forecasting

Recently, many deep learning models have been widely used in the field of multivariate time series forecasting. While traditional approaches like ARIMA [1] dominated early research, the field has witnessed a paradigm shift toward neural architectures capable of modeling complex temporal patterns. Recurrent Neural Networks (RNNs) [20,21] were initially explored as promising solutions. After the rise of attention mechanisms, Attention-based RNNs [22,23] introduced temporal attention to explore the long-range dependencies for forecasting. Transformers can capture temporal dependencies and inter-variable correlations due to their attention mechanism, making them an ideal choice for time series modeling tasks. Informer [8] introduced a ProbSparse self-attention mechanism to efficiently extract the most significant keys. PatchTST [9] proposed a patching strategy to allow the model to see the longer historical sequence. Although the Transformer has natural advantages in learning semantic correlations between sequence elements, the design of the Transformer is complex and cannot represent the key information of "order" in time series data. Dlinear [10] showed that a very simple linear model could outperform Transformer-based models on various common benchmarks, challenging the practicality of transformers in time series forecasting.

### 2.2   Adversarial attacks on time series forecasting

The vulnerability of time series forecasting models to adversarial attacks has emerged as a critical research frontier. Existing studies predominantly adapt attack paradigms from computer vision to temporal domains with adaptations in perturbation generation mechanisms. Pialla et al. [11] introduced an adversarial smooth perturbation approach by incorporating a smoothness penalty into the Basic Iterative Method (BIM) attack. Wu et al. [12] proposed a method for generating adversarial time series by introducing slight perturbations based on importance measurements. Mode et al. [13] utilized the BIM attack to target deep learning-based regression models. Xu et al. [14] developed a gradient-based technique to generate imperceptible perturbations. These studies aim to understand and address the weaknesses of time series forecasting models, but most of the design ideas for the attacks originate from the field of images, and there is a lack of research and exploration specific to time series data itself.

### 2.3   Frequency analysis on time series forecasting

There is an increasing trend of integrating frequency analysis into forecasting models. By transforming the input sequence into the frequency domain, the autocorrelation in the input can be effectively captured, thereby enhancing the forecasting performance of the models. FEDformer [15] proposed a method that combines the Transformer with seasonal trend decomposition, taking advantage of the fact that most time series tend to have sparse representations on well-known bases such as the Fourier transform. FourierGNN [16] rethought multivariate time series forecasting from a purely graphical perspective, considering each sequence value as a graph node and performing matrix multiplication in Fourier space. FreTS [17] converted the time-domain signals into frequency-domain complexes and learned the real and imaginary parts of the frequency components using MLP. These studies emphasize the significance of auto-correlation and frequency analysis in advanced time series modeling.

## 3   Method

### 3.1   Targeted Attacks on Time Series Forecasting

Given a time series $\vec{x} = [\mathbf{x}_1, \mathbf{x}_2 \ldots, \mathbf{x}_L]$, where each $\mathbf{x}_t \in \mathbb{R}^{M \times L}$ at time step $t$ is a vector of dimension M, a time series forecasting task aims to forecast T future values $[\mathbf{x}_{L+1}, \ldots, \mathbf{x}_{L+T}]$.

Given a time series forecasting model $\mathbf{f}$ and a time series $\vec{x}$, an adversarial attack aims to perturb the forecasting model by adding a small perturbation $\vec{r}$ to $\vec{x}$. We call the perturbed time series adversarial example, which can be expressed as $\vec{x}_{adv} = \vec{x} + \vec{r}$. The adversarial example $\vec{x}_{adv}$ is intended to worsen the output of the forecasting model.

In a classification problem, since there is a clear decision boundary, the attacker only needs to apply the smallest level of perturbation required to cause the model to misclassify. Unlike the classification task, sequence prediction is usually evaluated in terms of the Mean Absolute Error (MAE) and Mean Squared Error (MSE) between the model's predicted sequence and the ground-truth sequence. This allows even small perturbations in adversarial samples to affect the model's predictions. If the perturbations are too small, they may not be able to produce substantial errors in the model output. Conversely, larger perturbations may affect the stealthiness of the adversarial sample, making it easily detected, which requires the attack algorithm to constrain the magnitude of the perturbation directly. In a targeted attack, the attacker's goal is to make the model output as close as possible to the predetermined target sequence. Thus, the perturbation is formulated as a constrained optimization problem:

$$\min_{\vec{r}} \|f(\vec{x}+\vec{r}) - \vec{y}'\|_2, s.t. \|\vec{r}\|_2 < \varepsilon \tag{1}$$

where $\vec{y}'$ is the target sequence. In this paper, we focus on targeted attacks; therefore, only algorithmic formulations focusing on targeted attacks are presented.

Current white-box time series prediction attacks use the FSGM [5] algorithm and its modification PGD [6]. FSGM adds perturbations to the sequence directly using gradient notation:

$$\vec{x}_{adv} = \vec{x} - \varepsilon \cdot sign(\nabla_{\vec{x}} \|f(\vec{x}+\vec{r}) - \vec{y}'\|) \tag{2}$$

PGD decomposes the FGSM into an iterative process while projecting the adversarial samples back into the neighborhood sphere of the real samples as a constraint after each iteration step:

$$x_{adv}^n = \Pi_\epsilon \left( x_{adv}^{n-1} - \varepsilon \cdot \nabla_x \|f(x_{adv}^{n-1}) - \vec{y}'\| \right) \tag{3}$$

### 3.2 Carlini and Wagner's attack on TSF

Carlini and Wagner's adversarial attack (C&W) [18] was a state-of-the-art adversarial attack for image classifiers and has quickly been used in time series classification tasks. C&W is an optimization-based method that aims to minimize:

$$\min \|r\|_2 + c \times \phi(x+r) \quad \text{with} \quad \phi(x+r) \leq 0 \tag{4}$$

where $c$ is a hyper-parameter balancing the trade-off between the L2 regularization and $\phi$, a function that enforces the misclassification. Due to its direct optimization of the target loss rather than relying on gradient symbols, this method typically outperforms traditional first-order adversarial attacks in terms of both attack performance and stability. Many improvements have been proposed for the function $\phi$, including adjustments to improve its performance on specific classification tasks or to enhance computational efficiency. However, these adaptations are not directly applicable to the prediction task, as forecasting models require the manipulation of continuous sequence outputs rather than discrete class labels.

To bridge this gap, we adapted the attack to time series forecasting by redefining the optimization objective to suit the prediction context. Specifically, we replace the function $\phi$ with the mean squared error (MSE) between the model's output for the adversarial samples and the target sequence. The original work applied the constraint $r \in [0,1]^n$. However, it is not well-suited for time-series prediction tasks with low dimensions and degrees of freedom. Here, we refer to the constraint strategy of the PGD algorithm, replacing this constraint with the projection of adversarial samples within the neighborhood of real samples. Therefore, the perturbation will be formulated by minimizing the following equation:

$$L^{(\text{tmp})} = \|\vec{r}\|_2 + \text{MSE}(f(\vec{x}+\vec{r}), \vec{y}') \tag{5}$$

### 3.3  Fourier transform and frequency domain loss

It has been shown that existing time series forecasting models suffer from frequency domain bias, especially in multidimensional time series data. In some complex scenarios, models tend to learn low-frequency features in the data and overlook high-frequency features. This bias prevents models from accurately capturing important high-frequency data features because they focus excessively on frequencies with higher energy. This inspires us to optimize adversarial samples through frequency domain analysis. To achieve this goal, a feasible idea is to transform the time series into the frequency domain. We hope to generate adversarial perturbations by optimizing the loss in the frequency domain.

The Discrete Fourier Transform (DFT) is an intuitive and effective method. It describes the label sequence as a linear combination of orthogonal time patterns by projecting the sequence onto a set of orthogonal sine and cosine bases, effectively bypassing the autocorrelation in the time domain. The DFT of a sequence $\vec{Y} = [y_0, y_1 \dots, y_{L-1}]$ is defined as its projection onto a set of orthogonal Fourier bases with different frequencies. The projection on the basis associated with frequency $k$ is computed as:

$$F_k = \sum_{t=0}^{T-1} y_t \exp\left(-j\left(\frac{2\pi k}{T}\right)t\right) \tag{6}$$

where $j$ is the imaginary unit, $\exp(\cdot)$ is the Fourier basis orthogonal for different $k$ values. DFT refers to the set of projections $F = [F_1, \dots, F_{L-1}]$, which can be computed via the fast Fourier transform (FFT) algorithm.

Since the time series model is normalized in the time domain during the training process as well as before the adversarial attack, to ensure that the frequency domain loss has a close order of magnitude to the time domain loss, after the DFT we normalize the obtained sequence in the frequency domain, and denote the normalized spectral sequence as $F = \mathcal{F}(\vec{Y})$. The constrained optimization problem in the time domain is calculated according to (7). Since FFT has been proven to be differentiable, the frequency domain loss can be optimized using standard stochastic gradient descent methods. Similarly, using DFT to transform the prediction of adversarial samples and the target label sequence into the frequency domain, the loss in the frequency domain is calculated as follows:

$$L^{(\text{feq})} = |\mathcal{F}(\vec{x}+\vec{r}) - \mathcal{F}(\vec{x})| + |\mathcal{F}(f(\vec{x}+\vec{r})) - \mathcal{F}(f(\vec{x}))| \tag{7}$$

Considering the numerical characteristics of the label sequence, we choose to use L1 loss instead of squared loss in the frequency domain. Different frequency components typically exhibit significant amplitude differences. Low frequencies exhibit significantly higher amplitudes compared to high frequencies, leading to potential instability with square loss. By utilizing L1 loss, we achieve a more balanced and stable optimization process.

Finally, the loss of constrained optimization in the time and frequency domains are fused as follows, where $0 \leq \alpha \leq 1$ controls the relative strength of frequency-domain loss:

$$L = \alpha \cdot L^{(\text{feq})} + (1 - \alpha) \cdot L^{(\text{tmp})} \tag{8}$$

The overall attack procedure of Fre-CW is summarized in Algorithm 1.

---

**Algorithm 1** Attack Procedure of Fre-CW

**Input**: Time series forecasting model $f$, input $x$, target sequence $y'$, iterations $N_{iter}$, hyper-parameter $\alpha$, perturbation constraint $\varepsilon$
**Output**: adversarial sample $x_{\text{adv}}$
1: **Initialize** $x_{\text{adv}} = x$
2: for $i$ in range ($N_{iter}$) do
3:     $L^{(\text{tmp})} = \| x_{\text{adv}} - x \|_2 + \text{MSE}(f(x_{\text{adv}}), y')$
4:     Calculate $\mathcal{F}(x_{\text{adv}})$, $\mathcal{F}(x)$, $\mathcal{F}(f(x_{\text{adv}}))$ and $\mathcal{F}(y')$
5:     # use DFT and normalize the spectral sequence
6:     $L^{(\text{feq})} = |\mathcal{F}(x_{adv}) - \mathcal{F}(x)| + |\mathcal{F}(f(xadv)) - \mathcal{F}(y')|$
7:     $L = \alpha \cdot L^{(\text{feq})} + (1 - \alpha) \cdot L^{(\text{tmp})}$
8:     $x_{\text{adv}} \leftarrow \arg\min L$
9:     $x_{\text{adv}} = \text{Clip}_{x,\varepsilon}(x_{\text{adv}})$
10: **end for**
11: **return** $x_{\text{adv}}$

---

## 4 Experiment

### 4.1 Experimental setup

**Datasets and target models.** We evaluate the performance of our proposed C&W and Fre-CW on 6 popular time series forecasting datasets, including Weather, Electricity (ECL), and 4 ETT datasets (ETTh1, ETTh2, ETTm1, ETTm2). These datasets have been extensively utilized for benchmarking and are publicly available on [19]. We evaluated our method on current mainstream time series prediction models, including PatchTST [9] based on Transformer and Dlinear [10] representing linear models. As in the original papers, all models follow the same experimental setup with prediction length ($T = 96$) for all datasets.

Table 1. Detailed dataset descriptions.

| Dataset | D | Train / validation / test | Frequency | Domain |
|---|---|---|---|---|
| ETTh1 | 7 | 8545 / 2881 / 2881 | Hourly | Health |
| ETTh2 | 7 | 8545 / 2881 / 2881 | Hourly | Health |
| ETTm1 | 7 | 34465 / 11521 / 11521 | 15min | Health |
| ETTm2 | 7 | 34465 / 11521 / 11521 | 15min | Health |
| Weather | 21 | 36792 / 5271 / 10540 | 10min | Weather |
| ECL | 321 | 18317 / 2633 / 5261 | Hourly | Electricity |

D denotes the number of variates. Frequency denotes the sampling interval of time points.

**Baseline methods.** We choose FSGM [5] and PGD [6] algorithms as our baselines. Both algorithms have been studied in time series prediction and can be used for targeted attacks.

**Metrics and implementation details.** The MAE and MSE between the prediction results of adversarial samples and the target sequence are used as evaluation indicators. For the target sequence, we borrowed fixed perturbation measurements from the time series classification domain and assigned a fixed perturbation threshold to the normalized sequence of the entire time series. In this work, the target sequence is set as $\vec{y}' = \vec{y} + \alpha \cdot |\vec{y}|$ and $\alpha = 0.1$. For all attack algorithms, the perturbation threshold for adversarial samples $\varepsilon = 0.25$. We select the hyper-parameter $\alpha$ from [0.2,0.4,0.6,0.8]. We use Adam to optimize the adversarial samples in C&W and Fre-CW, with a learning rate set to 0.01 and a maximum iteration count set to 100.

### 4.2  Main results

We present the attack performance of our method and baseline in Table 2 and Table 3, respectively. Overall, Fre-CW greatly improves the performance of time series forecasting of targeted attacks on both models. C&W and Fre-CW, which directly optimize the adversarial samples, perform better than FSGM and PGD, which are based on gradient sign descent. Fre-CW demonstrates superior targeted attack performance, achieving MAE of 0.340, 0.250, 0.198, 0.169, 0.086 and 0.154 on these datasets against PatchTST. On PatchTST for the ETTm1 dataset, Fre-CW reduces the MAE by 0.009 compared to CW, an improvement comparable to the difference between PGD and C\&W, and reduces the MSE by 0.014. Similar gains are evident in the other datasets, which can be attributed to the role of the frequency-domain loss in optimizing the attack samples, validating the effectiveness of Fre-CW.

Table 2. Targeted attack performance on PatchTST.

| Attack | No Attack | FGSM | PGD | C&W | Fre-CW |
|---|---|---|---|---|---|
| Metric | MAE MSE | MAE MSE | MAE MSE | MAE MSE | MAE MSE |
| ETTh1 | 0.420 0.386 | 0.359 0.308 | 0.357 0.297 | <u>0.349 0.285</u> | **0.340 0.271** |
| ETTh2 | 0.357 0.270 | 0.294 0.208 | 0.294 0.189 | <u>0.263 0.166</u> | **0.250 0.156** |
| ETTm1 | 0.358 0.300 | 0.264 0.196 | 0.262 0.171 | <u>0.210 0.125</u> | **0.198 0.117** |
| ETTm2 | 0.280 0.171 | 0.229 0.132 | 0.221 0.116 | <u>0.185 0.095</u> | **0.169 0.088** |
| Weather | 0.216 0.169 | 0.196 0.130 | 0.125 0.111 | <u>0.093 0.082</u> | **0.086 0.080** |
| ECL | 0.247 0.150 | 0.187 0.103 | 0.185 0.092 | <u>0.164 0.077</u> | **0.154 0.071** |

The best results are in **bold** and the second best are <u>underlined</u>.

Table 3. Targeted attack performance on Dlinear.

| Attack | No Attack | FGSM | PGD | C&W | Fre-CW |
|---|---|---|---|---|---|
| Metric | MAE MSE | MAE MSE | MAE MSE | MAE MSE | MAE MSE |
| ETTh1 | 0.403 0.371 | 0.342 0.308 | 0.336 0.279 | <u>0.331 0.270</u> | **0.316 0.254** |
| ETTh2 | 0.345 0.262 | 0.278 0.192 | 0.263 0.184 | <u>0.259 0.158</u> | **0.244 0.145** |
| ETTm1 | 0.358 0.308 | 0.282 0.229 | 0.267 0.196 | <u>0.239 0.169</u> | **0.228 0.161** |
| ETTm2 | 0.266 0.161 | 0.224 0.114 | 0.185 0.103 | <u>0.163 0.073</u> | **0.154 0.068** |
| Weather | 0.221 0.191 | 0.210 0.146 | 0.151 0.144 | <u>0.121 0.112</u> | **0.120 0.111** |
| ECL | 0.255 0.159 | 0.204 0.101 | 0.175 0.110 | <u>0.159 0.082</u> | **0.151 0.077** |

The best results are in **bold** and the second best are <u>underlined</u>.

We visualized a sequence in the Etth2 dataset using CW and Fre-CW, including the time-domain sequence and the first 30 major frequencies. The adversarial samples generated by the two attack algorithms and their major frequencies are illustrated in Fig. 2. The adversarial samples of Fre-CW have fewer burrs in the time domain, which suggests that the adversarial samples generated by Fre-CW have a better stealthy performance. The model outputs of the two adversarial samples and their principal frequencies are illustrated in Fig. 3 and the output sequence of the Fre-CW adversarial sample is closer to the target sequence in the key peaks in the frequency domain, indicating its better attack performance.

### 4.3    Non-target attack

We performed non-target attack experiments for the PatchTST model on Etth1 and Etth2. The results are displayed in Table 4. The goal of the non-target attack is to maximize the loss of the output sequence and the target sequence (here we use the true sequence $\vec{y}$. Thus, the larger the metrics in Table 4, the better the attack performance. Both C&W and Fre-CW outperform the baseline algorithm, and Fre-CW shows better than C&W. This demonstrates that the improved C&W and Fre-CW can be used for the non-target attack with excellent attack performance.

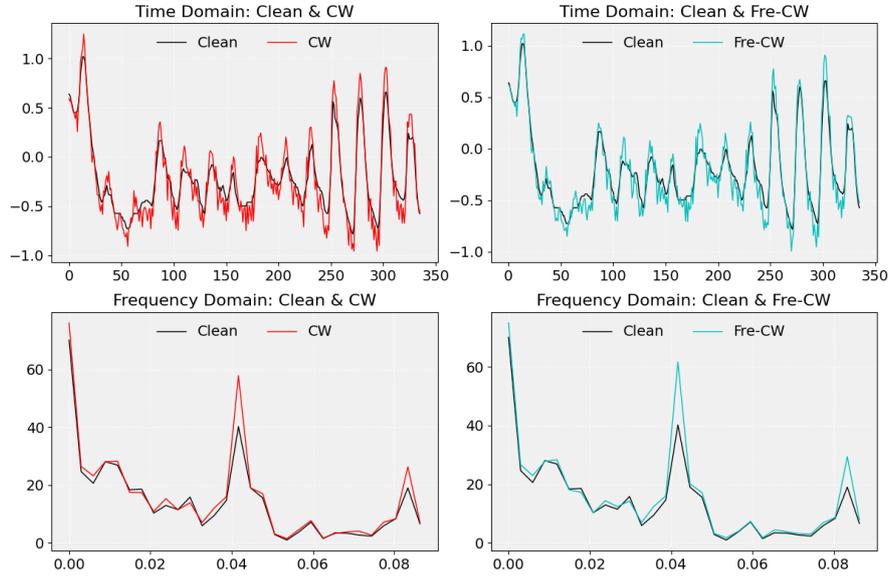

**Fig. 2.** Visualization of C&W and Fre-CW generated adversarial samples in time and frequency domains.

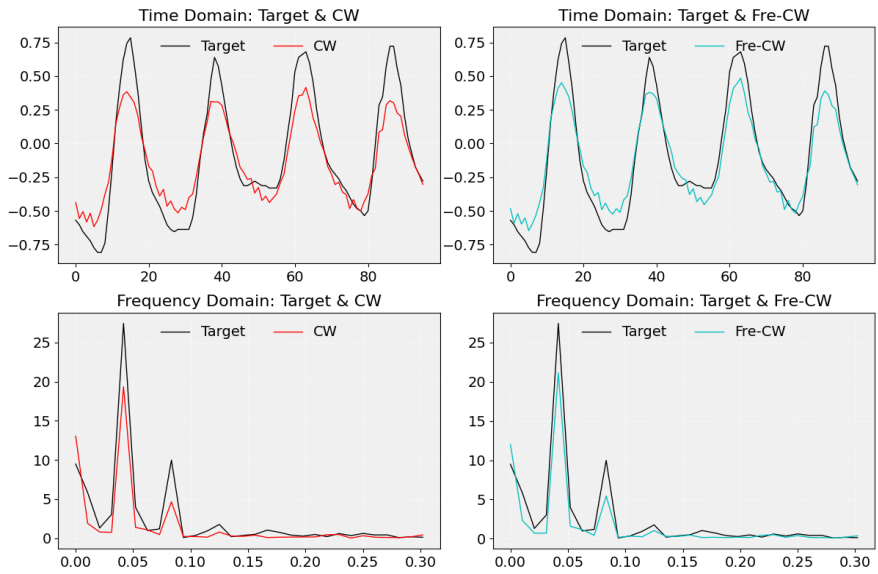

**Fig. 3.** Visualization of the output sequence after C&W and Fre-CW attacks in the time and frequency domains.

Table 4. Non-targeted attack performance.

| Attack | No Attack | FGSM | PGD | C&W | Fre-CW |
|---|---|---|---|---|---|
| Metric | MAE MSE | MAE MSE | MAE MSE | MAE MSE | MAE MSE |
| ETTh1 | 0.405 0.382 | 0.490 0.507 | 0.498 0.513 | <u>0.517 0.520</u> | **0.531 0.574** |
| ETTh2 | 0.336 0.274 | 0.488 0.438 | 0.514 0.461 | <u>0.534 0.499</u> | **0.552 0.523** |

The best results are in **bold** and the second best are <u>underlined</u>.

### 4.4  Hyperparameter sensitivity

The key hyperparameter of Fre-CW is the frequency loss strength $\alpha$. Fig.4 summarizes the performance for different $\alpha$ on the Etth1 dataset. Overall, increasing $\alpha$ from 0 to 1 increases and then decreases attack performance. This loss reduction trend seems consistent across different datasets, supporting the role of frequency domain loss in adversarial attacks. It is worth noting that the best attack performance typically occurs close to an $\alpha$ value between 0.5 and 0.7, although the minimum take-off points for MAE and MSE are slightly different. What is certain is that combining signal characteristics in both the time and frequency domains can improve attack performance.

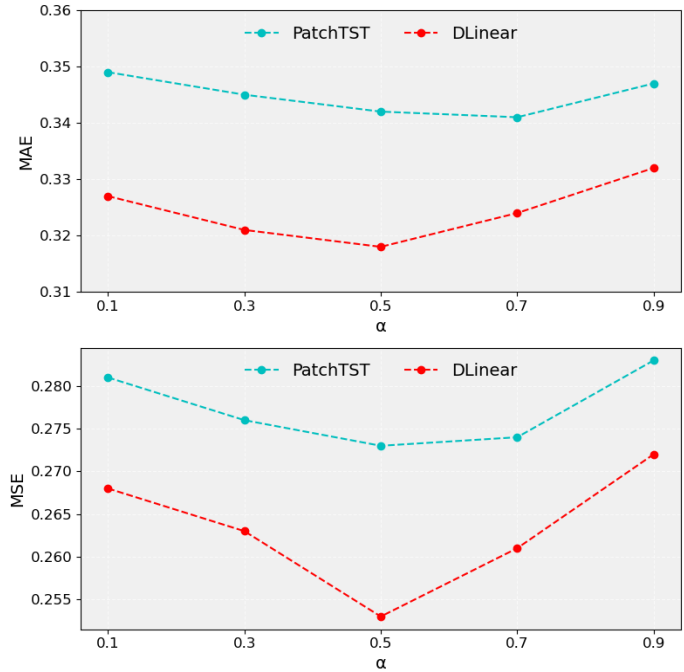

**Fig. 4.** Varying strength of frequency loss ($\alpha$) results on Etth1.

Another parameter that affects the performance of the attack is the perturbation constraints $\varepsilon$. Fig. 5 illustrates the variation curves of different attack methods with the growth of $\varepsilon$. Larger perturbation constraints naturally result in a broader adversarial sample space, leading to better attack performance. Notably, the disparity in performance between the Fre-CW and C&W algorithms initially increases and subsequently decreases as the perturbation constraint expands. This phenomenon can be attributed to the following: under small perturbation constraints, the optimization of high-frequency components in the frequency domain remains insufficient, thereby limiting the advantages of the Fre-CW algorithm. Conversely, when the perturbation constraints are large, both algorithms approach saturation and converge within a region of proximity.

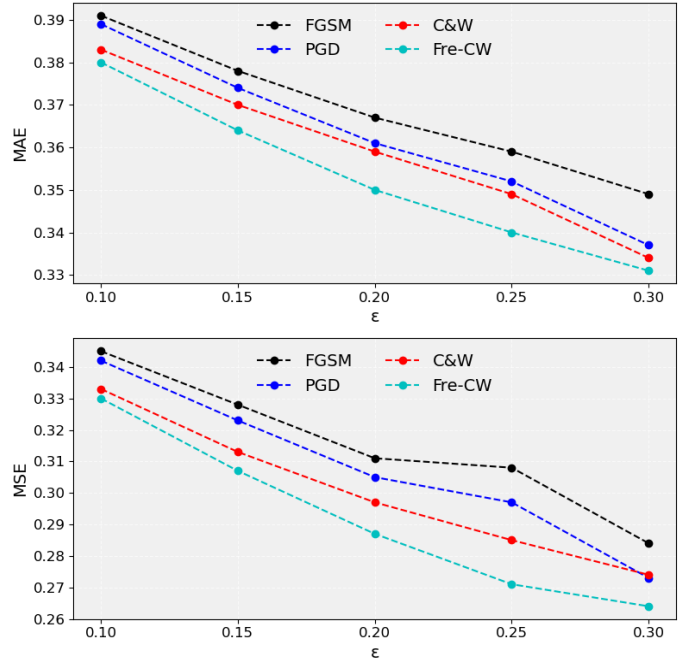

**Fig. 5.** Different $\varepsilon$ values on Etth1.

## 5      Conclusion

In this study, we investigate targeted adversarial attacks under the task of time series forecasting. We present an enhanced version of the C&W algorithm for time series prediction, incorporating frequency domain features to create the Fre-CW method. It transforms the labeled sequences into the frequency domain and optimizes the adversarial samples using time and frequency domain loss. Our experiments demonstrate that Fre-CW enhances attack performance and stealth while showcasing adaptability to mainstream forecasting models. These findings highlight the potential of frequency-aware adversarial strategies in exposing vulnerabilities in time series forecasting models.

Although our work provides new ideas and a good starting point for future research in this area, many challenges remain to be explored in time series adversarial attacks. In this work, we use the Fourier transform for domain transformation. The Fourier transform assumes that each label is sampled regularly, limiting its applicability in irregular time series. Future use of advanced domain transformation techniques, such as the non-uniform discrete Fourier transform, could extend Fre-CW to handle irregular time series. In addition, timestamp-based time series prediction models are beginning to be widely studied, and timestamp information can be introduced into adversarial research to further improve the perturbation realism and attack performance in the future.

## References

1. P. Mondal, L. Shit, S. Goswami, "Study of effectiveness of time series modeling (ARIMA) in forecasting stock prices," Int. J. Comput. Sci. Eng. Appl., vol. 4, no. 2, pp. 13, 2014.
2. C. Deb, F. Zhang, J. Yang, S.E. Lee, K.W. Shah, "A review on time series forecasting techniques for building energy consumption," Renewable Sustainable Energy Rev., vol. 74, pp. 902–924, 2017.
3. R.G. Cirstea, C. Guo, B.Y. "Graph attention recurrent neural networks for correlated time series forecasting," in Proc. KDD MiLeTS19, 2019.
4. I.J. Goodfellow, J. Shlens, C. Szegedy, "Explaining and harnessing adversarial examples," arXiv preprint arXiv:1412.6572, 2014.
5. Z. Chen, K. Dost, X. Zhu, X. Chang, G. Dobbie, J. Wicker, "Targeted Attacks on Time Series Forecasting," in Pacific-Asia Conf. Knowl. Discov. Data Min., pp. 314-327, Cham: Springer Nature Switzerland, 2023.
6. Y. Govindarajulu, A. Amballa, P. Kulkarni, M. Parmar, "Targeted Attacks on Timeseries Forecasting," arXiv preprint arXiv:2301.11544, 2023.
7. X. Piao, X. Zheng, C. Chen, T. Murayama, Y. Matsubara, and Y. Sakurai, "Fredformer: Frequency debiased transformer for time series forecasting," Proc. 30th ACM SIGKDD Conf. Knowledge Discovery Data Mining, pp. 2400–2410, 2024.
8. H. Zhou, S. Zhang, J. Peng, S. Zhang, J. Li, H. Xiong, and W. Zhang, "Informer: Beyond efficient transformer for long sequence time-series forecasting," Proc. AAAI Conf. Artif. Intell., vol. 35, pp. 11106–11115, 2021.
9. Y. Nie, N. H. Nguyen, P. Sinthong, and J. Kalagnanam, "A time series is worth 64 words: Long-term forecasting with transformers," Proc. Int. Conf. Learn. Represent., 2023.
10. A. Zeng, M. Chen, L. Zhang, and Q. Xu, "Are transformers effective for time series forecasting?" Proc. AAAI Conf. Artif. Intell., vol. 37, no. 9, pp. 11121–11128, 2023.
11. G. Pialla, H. I. Fawaz, M. Devanne, J. Weber, L. Idoumghar, P.-A. Muller, C. Bergmeir, D. F. Schmidt, G. I. Webb, and G. Forestier, "Time series adversarial attacks: An investigation of smooth perturbations and defense approaches," Int. J. Data Sci. Anal., pp. 1–11, 2023.
12. T. Wu, X. Wang, S. Qiao, X. Xian, Y. Liu, and L. Zhang, "Small perturbations are enough: Adversarial attacks on time series prediction," Inf. Sci., vol. 587, pp. 794–812, 2022.

13. G. R. Mode and K. A. Hoque, "Adversarial examples in deep learning for multivariate time series regression," in Proc. IEEE Appl. Imagery Pattern Recognit. Workshop (AIPR), 2020, pp. 1–10.
14. A. Xu, X. Wang, Y. Zhang, T. Wu, and X. Xian, "Adversarial attacks on deep neural networks for time series prediction," in Proc. 10th Int. Conf. Internet Comput. Sci. Eng., 2021, pp. 8–14.
15. T. Zhou, Z. Ma, Q. Wen, X. Wang, L. Sun, and R. Jin, "FEDformer: Frequency enhanced decomposed transformer for long-term series forecasting," in Proc. Int. Conf. Mach. Learn., 2022.
16. K. Yi, Q. Zhang, W. Fan, H. He, L. Hu, P. Wang, N. An, L. Cao, and Z. Niu, "Fouriergnn: Rethinking multivariate time series forecasting from a pure graph perspective," in Proc. Adv. Neural Inf. Process. Syst., 2023a.
17. K. Yi, Q. Zhang, W. Fan, S. Wang, P. Wang, H. He, N. An, D. Lian, L. Cao, and Z. Niu, "Frequency-domain mlps are more effective learners in time series forecasting," in Proc. Adv. Neural Inf. Process. Syst., 2023b.
18. N. Carlini and D. Wagner, "Towards evaluating the robustness of neural networks," in Proc. 2017 IEEE Symp. Security Privacy, San Jose, CA, USA, 2017, pp. 39–57.
19. H. Wu, J. Xu, J. Wang, and M. Long, "Autoformer: Decomposition transformers with Auto-Correlation for long-term series forecasting," in Adv. Neural Inf. Process. Syst., 2021.
20. R. Wen, K. Torkkola, B. Narayanaswamy, and D. Madeka, "A multi-horizon quantile recurrent forecaster," in Adv. Neural Inf. Process. Syst., 2017.
21. R. Yu, S. Zheng, A. Anandkumar, and Y. Yue, "Long-term forecasting using tensor-train RNNs," arXiv preprint arXiv:1711.00073, 2017.
22. S.-Y. Shih, F.-K. Sun, and H.-Y. Lee, "Temporal pattern attention for multivariate time series forecasting," Mach. Learn., 2019.
23. H. Song, D. Rajan, J. Thiagarajan, and A. Spanias, "Attend and diagnose: Clinical time series analysis using attention models," in Proc. AAAI Conf., 2018.